\documentclass[10pt,twocolumn,letterpaper]{article}

\usepackage{cvpr}
\usepackage{times}
\usepackage{epsfig}
\usepackage{graphicx}
\usepackage{amsmath}
\usepackage{amssymb}
\usepackage{booktabs}
\usepackage{threeparttable}
\usepackage{xspace}
\usepackage[toc,page]{appendix}

\usepackage[breaklinks=true,bookmarks=false]{hyperref}

\cvprfinalcopy 

\def\cvprPaperID{2812} 

\ifcvprfinal\fi
\begin{document}

\title{Discriminative Learning of Latent Features for Zero-Shot Recognition}

\author{Yan Li$^{\,1,2}$, Junge Zhang$^{\,1,2}$, Jianguo Zhang$^{\,3}$, Kaiqi Huang$^{\,1,2,4}$\\
$^{1}$ CRIPAC \& NLPR, CASIA  $^{2}$ University of Chinese Academy of Sciences\\
$^{3}$ Computing, School of Science and Engineering, University of Dundee, UK\\
$^{4}$ CAS Center for Excellence in Brain Science and Intelligence Technology\\
{\tt\small yan.li@cripac.ia.ac.cn, jgzhang@nlpr.ia.ac.cn, j.n.zhang@dundee.ac.uk, kqhuang@nlpr.ia.ac.cn}
}

\maketitle

\begin{abstract}
Zero-shot learning (ZSL) aims to recognize unseen image categories by learning an embedding space between image and semantic representations. For years, among existing works, it has been the center task to learn the proper mapping matrices aligning the visual and semantic space, whilst the importance to learn discriminative representations for ZSL is ignored. In this work, we retrospect existing methods and demonstrate the necessity to learn discriminative representations for both visual and semantic instances of ZSL. We propose an end-to-end network that is capable of 1) automatically discovering discriminative regions by a zoom network; and 2) learning discriminative semantic representations in an augmented space introduced for both user-defined and latent attributes.  Our proposed method is tested extensively on two challenging ZSL datasets, and the experiment results show that the proposed method significantly outperforms state-of-the-art methods.
\end{abstract}

\section{Introduction}
\label{section:intro}

In recent years, zero-shot learning (ZSL) has gained its popularity in object recognition task \cite{akata2016label,fu2014transductive,fu2015transductive,huang2016local,karessli2017gaze,lampert2009learning,li2017zero,xian2017zero}. Unlike traditional object recognition methods that seek to predict the presence of an object instance by assigning its image label as one of the categories \textit{seen} in the training set, zero-shot learning aims to recognize an object instance from a new category \textit{never seen} before. Therefore, in the ZSL task, the seen categories in the training set and the unseen categories in the test set are disjoint. Typically, the descriptors of categories (\eg user-defined attribute annotations \cite{akata2016label,lampert2009learning}, the text descriptions of the categories \cite{reed2016learning}, the word vectors of the class names \cite{frome2013devise,norouzi2013zero}, \etc) are provided for both seen and unseen classes; some of those descriptors are shared between categories. Those descriptors are often called \textit{side information} or \textit{semantic} representations. In this work, we focus on learning for ZSL with attributes.

\begin{figure}[]
    \begin{center}
        \includegraphics[width=8cm]{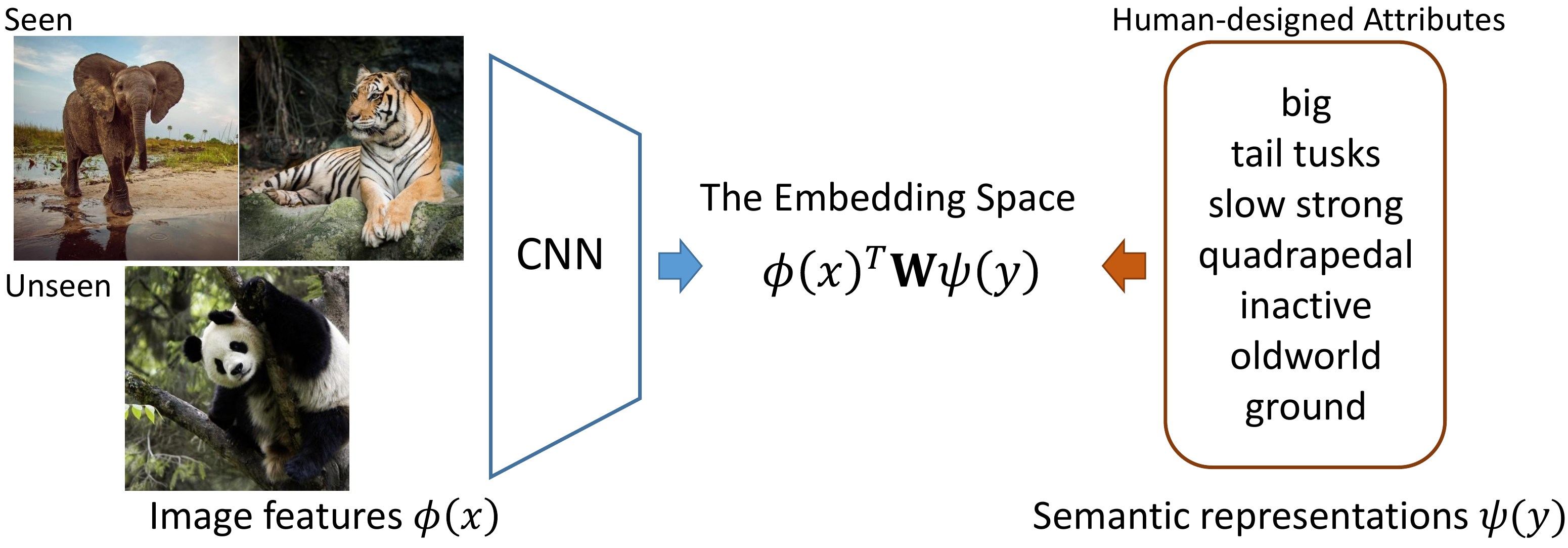}
    \end{center}
    \vspace{-1em}
        \caption{The typical ZSL approaches aim to find an embedding space where the image features $\phi(x)$ and semantic representations $\psi(y)$ are embedded.}
    \label{fig:emb}
    \vspace{-1.5em}
    \end{figure}

As shown in Figure \ref{fig:emb}, a general assumption under the typical ZSL methods is that there exists a shared embedding space, in which a mapping function, $F(x,y;\mathbf{W}) = \phi(x)^T\mathbf{W}\psi(y)$, is defined to measure the \textit{compatibility} between the image features $\phi(x)$ and the semantic representations $\psi(y)$ for both seen and unseen classes. $\mathbf{W}$ is the visual-semantic mapping matrix to be learned. Existing approaches of ZSL mainly focus on introducing linear or non-linear modelling methods, utilizing various optimization objectives and designing different specific regularization terms to learn the visual-semantic mapping, more specially, to learn $\mathbf{W}$ for ZSL.

To date, the learning of the mapping matrix $\mathbf{W}$, though important to ZSL, is mainly driven by minimizing the alignment loss between the visual and semantic space. However, the final goal of ZSL is to classify unseen categories. Therefore, the visual features $\phi(x)$ and semantic representations $\psi(y)$, should arguably be \textit{discriminative} to recognize different objects. Unfortunately, this issue has been thus far neglected in ZSL and almost all the methods follow the same paradigm: 1) extracting image features by hand-crafting or using pre-trained CNN models; and 2) utilizing the human-designed attributes as the semantic representations. There are some pitfalls existed in this paradigm.

Firstly, the image features $\phi(x)$ either crafted manually or from a pre-trained CNN model may be not representative enough for zero-short recognition task. Though the features from a pre-trained CNN model are learned, yet restricted to a fixed set of images (\eg, ImageNet \cite{russakovsky2015imagenet}), which is not optimal for a particular ZSL task.

Secondly, the user-defined attributes $\psi(y)$ are semantically descriptive, but they are not exhaustive, thus limiting its discriminativeness in classification. There may exist discriminative visual clues not reflected by the pre-defined attributes in ZSL datasets, \eg, the huge mouths of \textit{hippos}. On the other hand, as shown in Figure \ref{fig:emb}, the annotated attributes, such as \textit{big}, \textit{strong} and \textit{ground}, are shared in many object categories. This is desired for knowledge transfer between categories, especially from seen to unseen categories. However, if two categories (\eg \textit{cheetah} and \textit{tiger}) share too many (user-defined) attributes, they will be hardly distinguishable in the space of attribute vectors.

Thirdly, low-level feature extraction and embedding space construction in existing ZSL approaches are treated separately, and usually carried out in isolation. Therefore, few existing work ever considers those two components in a unified framework.

To address those pitfalls, we propose an end-to-end model capable of learning latent discriminative features (LDF) for ZSL in both visual and semantic space. Specifically, our contributions are:

1) A cascaded \textit{zooming} mechanism to learn features from object-centric regions. Our model can automatically identify the most discriminative region in an image and then zoom it into a larger scale for learning in a cascaded network structure. In this way, our model can concentrate on learning features from a region with object as a focus.

2) A framework to jointly learn the latent attributes and the user-defined attributes. We formulate the learning of latent attributes as a category-ranking problem to ensure the learned attributes are discriminative. Meanwhile, the discriminative region mining and the latent attributes modelling are jointly learned in our model and assist each other to gain further improvement.

3) An end-to-end network structure for ZSL. The obtained image features can be regulated to be more compatible with the semantic space, which contains both the user-defined attributes and latent discriminative attributes.

\begin{figure*}[]
    \begin{center}
        \includegraphics[height=6.27cm]{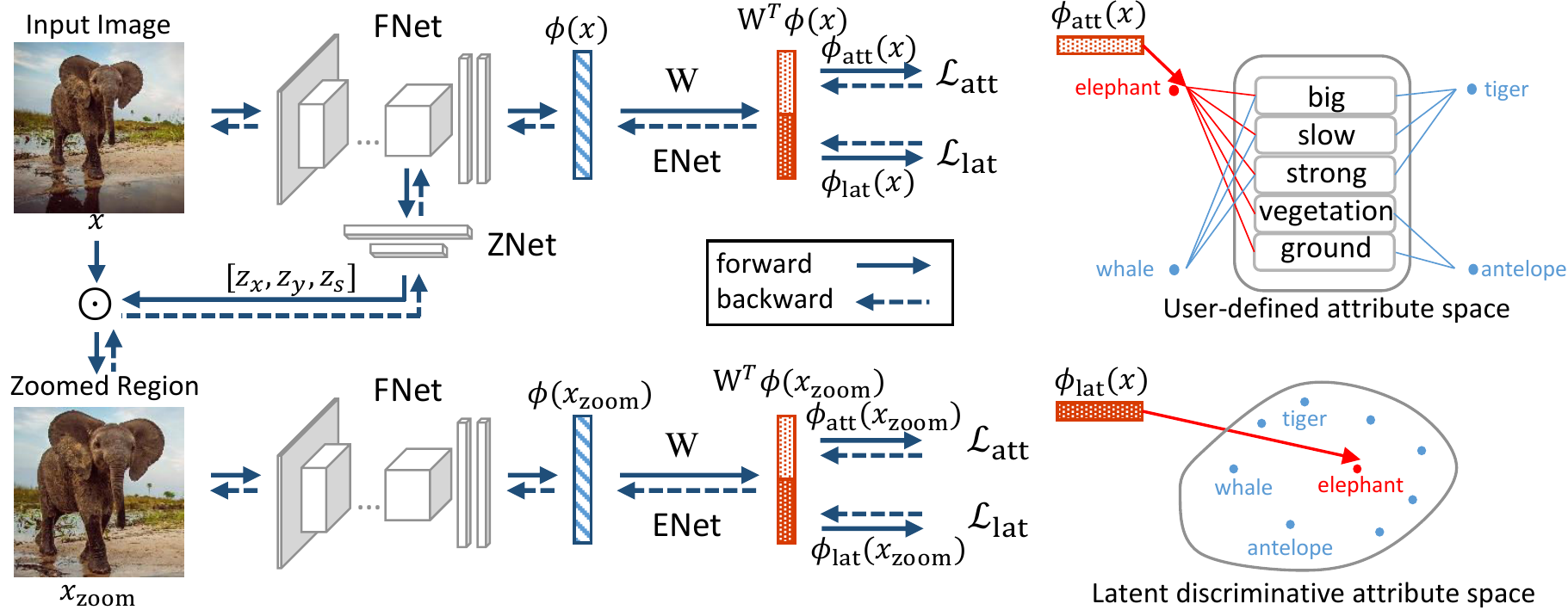}
    \end{center}
    \vspace{-1em}
        \caption{The framework of the proposed Latent Discriminative Features Learning (LDF) model. The coarse-to-fine image representations are projected into both user-defined attributes and latent attributes. The user-defined attributes are usually shared between different categories while the latent attributes are learned to be discriminative by regulating inter/intra class distances.}
    \label{fig:framework}
    \vspace{-1.5em}
    \end{figure*}

\section{Related Work}
Early works of zero-shot learning (ZSL) follow an intuitive way to object recognition that first trains different \textit{attribute classifiers} and then recognizes an image by comparing its predicted attributes with descriptions of unseen classes \cite{farhadi2009describing,lampert2009learning}. Among these works, Direct Attribute Prediction (DAP) model \cite{lampert2014attribute} predicts the posterior of each attribute, and then the class posteriors for an image are calculated by maximizing a posterior. Whilst in Indirect Attribute Prediction (IAP) \cite{lampert2014attribute} model, the attribute posteriors are computed from the class posterior of seen classes. In these methods, each attribute classifier is trained individually and the relationship between attributes for a class is not considered.
%

To address this issue, most of recent ZSL works are \textit{embedding-based} methods, which seek to build a common embedding space for images and their semantic features. The DeViSE model \cite{frome2013devise} and the ALE model \cite{akata2016label} are based on a bilinear embedding model, where a linear transformation matrix $\mathbf{W}$ is learned with a hinge ranking loss.
The ESZSL model \cite{romera2015embarrassingly} adds a Frobenius norm regularizer into the embedding space construction. The SJE model \cite{akata2015evaluation} combines several compatibility functions linearly to form a joint embedding space. The LatEM model \cite{xian2016latent} improves SJE with more nonlinearity by incorporating latent variables. Recently, the SCoRe model \cite{morgado2017semantically} adds a semantically consistent regularization to make the learned transformation matrix perform better on test images. The MFMR model \cite{xu2017matrix} learns the projection matrix by decomposing the visual feature matrix. The majority of ZSL methods thus far extract image features from whole image with fixed pre-trained CNN models. In contrast, image features in our model are learned to be more representative with the mining of latent discriminative regions and the end-to-end training style.

In typical embedding space construction approach, only the space of user-defined attributes is used to embed the seen and unseen classes. Different from this, the JSLA model \cite{peng2016joint,peng2017joint} and the LAD model \cite{jiang2017learning} propose to model latent attributes for ZSL, which are similar to our work. JSLA learns latent discriminative attributes by minimizing the intra class distance between the attributes; while in LAD the discriminativeness of latent attributes is indirectly achieved by training seen class classifiers over the latent attributes. Different from them, our model proposes to directly regulate both inter-class and intra-class distances between latent attributes to achieve the discriminativeness. What's more, JSLA and LAD still utilize the fixed pre-extracted image features, which are less representative than ours.

Another branch of ZSL approaches are based on \textit{hybrid models}, which aim to use the combination of seen classes to classify unseen images. The ConSE model \cite{norouzi2013zero} convexly combines the classification probabilities of seen classes to classify unseen objects. The SynC model \cite{changpinyo2016synthesized} introduces synthetic classifiers of unseen classes by linearly combining the classifiers of seen classes. In our method, when the learned latent attributes are utilized for ZSL prediction, the latent attribute prototype for an unseen class is obtained by combining the prototypes of seen classes. To this end, our prediction model is among the family of hybrid models; and beyond that our model also learns embeddings for both user-defined attributes and latent attributes in one network.

\section{Task Definition}

In the zero-shot learning task, the training set, \ie, the \textit{seen classes}, is defined as $\mathcal{S}\equiv \{(x_i^s,y_i^s)\}_{i=1}^{n_s}$, where $x_i^s\in\mathcal{X}_{\mathcal{S}}$ is the $i$-th image of the seen class and $y_i^s\in\mathcal{Y}_{\mathcal{S}}$ is its corresponding class label. The test set, \ie, the \textit{unseen classes}, is defined as $\mathcal{U}\equiv \{(x_j^u,y_j^u)\}_{j=1}^{n_u}$, where $x_j^u\in\mathcal{X}_{\mathcal{U}}$ denotes the $j$-th unseen image and $y_i^s\in\mathcal{Y}_{\mathcal{U}}$ is the label of it. The seen and unseen classes are disjoint, \ie, $\mathcal{Y}_{\mathcal{S}}\cap\mathcal{Y}_{\mathcal{U}}=\emptyset$. Additionally, the user-defined attributes for both seen and unseen classes can be denoted as $\mathcal{A}_\mathcal{S} \equiv \{\mathbf{a}_i^s\}_{i=1}^{c_s}$ and $\mathcal{A}_\mathcal{U} \equiv \{\mathbf{a}_j^u\}_{j=1}^{c_u}$, where $\mathbf{a}_i^s$ and $\mathbf{a}_j^u$ indicate the attribute vectors for the $i$-th seen class and the $j$-th unseen class, respectively. At the test stage, given a test image $x^u$ and the attribute annotations of test classes $\mathcal{A}_{\mathcal{U}}$, the goal of ZSL is to predict the corresponding category $y^u$ for $x^u$.

\section{Our Method}
The framework of the proposed method is illustrated in Figure \ref{fig:framework}. Note that the architecture in principle contains multiple scales and for clarity, we illustrate the network with two scales as an example. In each scale, the network consists of three different components, 1) the image feature network (FNet) to extract image representations, 2) the zoom network (ZNet) to locate the most discriminative region and then zoom it to larger scale and 3) the embedding network (ENet) to build the embedding space where the visual and semantic information are associated. For the first scale, the input of the FNet is the image of its original size and the ZNet is responsible for producing the zoomed region. Then for the second scale, the zoomed region is fed into the FNet to obtain more discriminative image features.


\subsection{The Image Feature Network (FNet)}
\label{section:FNet}
Different from existing works \cite{dinglow,morgado2017semantically,xu2017matrix}, we would like to learn image features together with embedding for zero-shot learning. Therefore, our framework starts with a compartment of convolutional nets responsible for learning image features, which is termed as FNet. The choice of the architecture of FNet is flexible; and two possible variants are considered in our approach, \ie, the VGG19 and the GoogLeNet. For VGG19, the FNet starts from conv1 to fc7; for GoogLeNet, it starts from conv1 to pool5. Given an image or a zoomed region $x$, the image representation is denoted as:
\begin{equation}
\phi(x)=\mathbf{W}_{\textrm{IF}}*x
 \end{equation}
 where $\mathbf{W}_{\textrm{IF}}$ indicates the overall parameters of the FNet, and $*$ denotes a set of operations of the FNet. Different from traditional ZSL approaches, the parameters of FNet are jointly trained with other parts in our framework; thus the obtained features are regulated well with the embedding component. We show that this leads to an performance improvement.

\subsection{The Zoom Network (ZNet)}
\label{section:ZNet}
The final goal of zero-shot learning is to classify different object categories. There exist studies showing that learning from object regions could benefit object categorization at image level \cite{fu2017look,zhang2007local}. Inspired by these studies, we hypothesize that there may exist some \textit{discriminative} regions in an image which benefit the zero-shot learning. Such a region could contain only object instance or object parts \cite{fu2017look}. On the other hand, for ZSL, a candidate region will also need to reflect the user-defined attributes, some of which describe the background, such as \textit{swim}, \textit{tree} and \textit{mountains}. Therefore, a target region is expected to contain some background to enhance the attributes embedding. We name this type of regions as \textit{object-centric} region. To identify them, we introduce the zoom network (ZNet) that adopts an incrementally \textit{zoom-in} approach to let the network automatically search a proper discriminative region from coarse to fine. The \textit{proper} in ZSL task means that the target region is discriminative for classification and meanwhile matched with the annotated attributes.

Specifically, our ZNet takes the output of the last convolutional layer in the FNet (\eg, conv5\_4 in VGG19) as the input. For computational efficiency, the candidate region is assumed as a square and its location can be represented with three parameters:
\begin{equation}
\label{equ:znet}
[z_x,z_y,z_s]=\mathbf{W}_{\textrm{Z}}*\phi(x)_{\text{conv}}
\end{equation}
where $z_x,z_y$ indicate the x-axis and y-axis coordinates for the center of the searched square, respectively, and $z_s$ represents the length of the square. The $\phi(x)_{\text{conv}}$ denotes the output of the last convolutional layer of the FNet. The ZNet is a two-stacked fully-connected layers (1024-3) followed by the sigmoid activation function and $\mathbf{W}_{\textrm{Z}}$ denotes the parameters of the ZNet.

After obtaining the location of the square, the searched region can be obtained by directly cropping from the original image. However, it is not convenient to optimize the non-continuous cropping operation in backward-propagation. Inspired by \cite{fu2017look}, the sigmoid function is utilized to first produce a two-dim continuous mask $\mathbf{M}(\text{x},\text{y})$. Formally,
\begin{equation}
\begin{aligned}
& \mathbf{M}_{\text{x}}=f(\text{x}-z_x+0.5z_s)-f(\text{x}-z_x-0.5z_s) \\
& \mathbf{M}_{\text{y}}=f(\text{y}-z_y+0.5z_s)-f(\text{y}-z_y-0.5z_s)
\end{aligned}
\end{equation}
where $f(x)=1/(1+\exp(-kx))$ and $k$ is set to 10 in all experiments.

Then the cropped region can be obtained by implementing element-wise multiplication $\odot$ between the original image $x$ and the continuous mask $\mathbf{M}$:
\begin{equation}
x^{\text{crop}} = x \odot \mathbf{M}
\end{equation}

Finally, to obtain better representation for finer localized cropped region, we further use the bilinear interpolation to adaptively zoom the cropped region to the same size with the original image.
The zoomed region is then fed into a copy of the FNet in the next scale to extract more discriminative representation.
\subsection{The Embedding Network (ENet)}
\subsubsection{The Baseline Embedding Model}
\label{section:BE}
The embedding network (ENet) aims to learn an embedding space where the visual and semantic information are associated. In this section, we first introduce a baseline embedding model, where the semantic representations, $\psi(y)$, is defined with the user-defined attributes $\mathcal{A}$. In this model, the mapping function to be learned is therefore defined as: $F(x,y;\mathbf{W})=\phi(x)^T\mathbf{W}\mathbf{a}^y$.

The attribute space $\mathcal{A}$ is adopted as the embedding space and the compatibility score is defined by the inner product:
\begin{equation}
\mathbf{s}=\langle \mathbf{W}^T\phi(x), \mathbf{a}^y\rangle
\end{equation}
where $\phi(x)$ is the $d$-dim image representation obtained by the FNet and $\mathbf{a}^y$ is the $k$-dim annotated attribute vector of category $y$. $\mathbf{W} \in \mathbb{R}^{d\times k}$ is the weight to learn in a fully connected layer, which can be considered as a linear project matrix that maps $\phi(x)$ to the attribute space $\mathcal{A}$.

The compatibility score measures the similarity between an image and the attribute annotations of classes. It is similar to the classification score in traditional object recognition task. Thus, to learn the matrix $\mathbf{W}$, a standard softmax loss can be used:
\begin{equation}
\label{equ:baseline}
  \mathcal{L} =-\frac{1}{N}\sum_{i}^{n}\mathrm{log}\frac{\mathrm{exp}(\mathbf{s})}{\sum_{c}\mathrm{exp}(\mathbf{s}^c)}, c\in \mathcal{Y_S}
  \end{equation}
\subsubsection{The Augmented Embedding Model}
\label{section:AE}
The baseline embedding model, adopted by most of existing ZSL methods, has achieved promising performance. However, it is based on user-defined attributes, which is of limited size, and usually not discriminative. To address this issue, we introduce an \textit{augmented} attribute space, where an image is projected into both user-defined attributes (UA) and \textit{latent} discriminative attributes (LA).

Specifically, our embedding network (ENet) learns a matrix $\mathbf{W}_{\textrm{aug}} \in \mathbb{R}^{d \times 2k}$ mapping the image features to a $2k$-dim augmented space, and the embedded image features $\phi_{\textrm{e}}(x)$ are computed as follows:
\begin{equation}
\label{equ:proj}
\phi_{\textrm{e}}(x) = \mathbf{W}_{\textrm{aug}}^{T}\phi(x), \quad \phi_{\textrm{e}}(x) \in \mathbb{R}^{2k}
\end{equation}

The goal is to associate the embedded image features $\phi_{\textrm{e}}(x)$ with both the UA and the LA. For simplicity, we equally divide $\phi_{\textrm{e}}(x)$ into two $k$-dim parts:
 \begin{equation}
 \label{equ:split}
 \phi_{\textrm{e}}(x)=[\phi_{\textrm{att}}(x); \phi_{\textrm{lat}}(x)], \quad \phi_{\textrm{att}}(x), \phi_{\textrm{lat}}(x) \in \mathbb{R}^{k}
 \end{equation}

 Then we let the first $k$-dim embedded feature $\phi_{\textrm{att}}(x)$ correspond to the UA and the second $k$-dim component $\phi_{\textrm{lat}}(x)$ being associated with the LA. Based on this assumption, for $\phi_{\textrm{att}}(x)$, similar to the baseline model, the softmax loss is utilized to train the ZSL model. Formally,
\begin{equation}
\label{equ:att_loss}
  \mathcal{L}_{\textrm{att}} =-\frac{1}{N}\sum_{i}^{n}\mathrm{log}\frac{\mathrm{exp}(\langle \phi_{\textrm{att}}(x), \mathbf{a} \rangle)}{\sum_{c}\mathrm{exp}(\langle \phi_{\textrm{att}}(x), \mathbf{a}^c \rangle)}, c\in \mathcal{Y_S}
  \end{equation}

For the second embedded feature $\phi_{\textrm{lat}}(x)$, the goal is to make the learned features be discriminative for object recognition. We propose to utilize the triplet loss \cite{weinberger2009distance} to learn the latent discriminative attributes with regulating the inter/intra class distances between latent attributes features:
\begin{equation}
\label{equ:lat_loss}
\mathcal{L}_{\textrm{lat}} = \max(0,m+d(\phi_{\textrm{lat}}(x_i),\phi_{\textrm{lat}}(x_k))-d(\phi_{\textrm{lat}}(x_i),\phi_{\textrm{lat}}(x_j)))
\end{equation}
where $x_i, x_k$ are images from the same class and $x_j$ is from a different class. $d(x,y)$ is the squared Euclidean distance between $x$ and $y$. $m$ is the margin of the triplet loss and is set to 1.0 for all experiments.

From (\ref{equ:proj}) and (\ref{equ:split}), it can be observed that the UA and LA features are mapped from the same image representation, but with two different matrices:
\begin{equation}
\label{equ:two_proj}
\begin{aligned}
& \phi_{\textrm{att}}(x) = \mathbf{W}_{\textrm{att}}^{T}\phi(x), \\
& \phi_{\textrm{lat}}(x) = \mathbf{W}_{\textrm{lat}}^{T}\phi(x), \quad [\mathbf{W}_{\textrm{att}};\mathbf{W}_{\textrm{lat}}]=\mathbf{W}_{\textrm{aug}}
\end{aligned}
\end{equation}

It is noted that $\mathbf{W}_{\textrm{att}}$ and $\mathbf{W}_{\textrm{lat}}$ are associated with different loss functions. $\phi_{\textrm{lat}}$ can be \textit{learned} to be discriminative by specifically exploiting the category information in (\ref{equ:lat_loss}).

For each scale, the network is trained with both the softmax loss and the triplet loss. For a two-scale network (\ie, $s1$ and $s2$), the whole LDF model is trained by the following loss function:
\begin{equation}
\mathcal{L} = \mathcal{L}_{\textrm{att}}^{s1} + \mathcal{L}_{\textrm{lat}}^{s1} + \mathcal{L}_{\textrm{att}}^{s2} + \mathcal{L}_{\textrm{lat}}^{s2}
\end{equation}
The final objective function for a multi-scale network could be constructed similarly by aggregating all the loss functions of all of scales.

\subsection{ZSL Prediction}
In the proposed LDF model, the test images can be projected into both user-defined attributes (UA) and latent attributes (LA) as in (\ref{equ:proj}). Thus, ZSL prediction can be performed in both the UA space and the LA space.

\noindent \textbf{Prediction with UA.} Given a test image $x$, it can be projected to the UA representation $\phi_\textrm{att}(x)$. To predict its class label, the compatibility scores can be used to select the most matched unseen categories:
\begin{equation}
\label{equ:UA}
  y^* = \underset{c\in\mathcal{Y_U}}{\arg\max} (\mathbf{s}^c) = \underset{c\in\mathcal{Y_U}}{\arg\max} \langle \phi_\textrm{att}(x), \textbf{a}^c\rangle
  \end{equation}

\noindent \textbf{Prediction with LA.} The test image $x$ can also be projected to the LA representation, $\phi_\textrm{lat}(x)$. To perform ZSL in the LA space, the LA prototypes for unseen classes are required.

Firstly, the LA prototypes for seen classes are computed. Concretely, all samples $x_i$ from the seen class $s$ are projected to their LA features and the mean of features are utilized as the LA prototype of class $s$, \ie, $\overline{\phi_\textrm{lat}^s} = \frac{1}{N} \sum_i \phi_\textrm{lat}(x_i)$.

Then, for an unseen class $u$, we compute the relationship between class $u$ and all the seen classes $\mathcal{S}$ in the UA space. This relationship can be obtained by solving the following ridge regression problem:
\begin{equation}
\label{equ:ridge}
\beta_c^u = \arg\min {\| \mathbf{a}^{u} - \sum{\beta_c^u \mathbf{a}^c} \|}_2^2 + \lambda{\| \beta_c^u \|}_2^2,  \quad c \in \mathcal{Y}_\mathcal{S}
\end{equation}

By applying the same relationship to the LA space, the prototype for unseen class $u$ can be obtained:
\begin{equation}
\label{equ:lat_prototype}
\overline{\phi_\textrm{lat}^u} = \sum\beta_c^u \overline{\phi_\textrm{lat}^c}, \quad c \in \mathcal{Y}_\mathcal{S}
\end{equation}

Finally, the classification result of a test image $x$ with LA representation $\phi_\textrm{lat}(x)$ can be achieved as following:
\begin{equation}
\label{equ:lat_prediction}
y^* = \underset{c\in\mathcal{Y_U}}{\arg\max} \langle \phi_\textrm{lat}(x), \overline{\phi_\textrm{lat}^c} \rangle
\end{equation}

\noindent \textbf{Combining multiple spaces.} We can consider both the UA and LA spaces and utilize the concated UA-LA feature $[\phi_\textrm{att}(x);\phi_\textrm{lat}(x)]$ to perform ZSL prediction. Formally,
\begin{equation}
\label{equ:prediction}
\begin{aligned}
y^* &=  \underset{c\in\mathcal{Y_U}}{\arg\max}(\langle [\phi_\textrm{att}(x);\phi_\textrm{lat}(x)],[\mathbf{a}^c;\overline{\phi_\textrm{lat}^c}] \rangle) \\
    &=  \underset{c\in\mathcal{Y_U}}{\arg\max}(\langle \phi_\textrm{att}(x), \mathbf{a}^c\rangle + \langle \phi_\textrm{lat}(x), \overline{\phi_\textrm{lat}^c} \rangle)
\end{aligned}
\end{equation}

\noindent \textbf{Combining multiple scales.} For a two-scale LDF model (\ie, $s1$ and $s2$). The UA and LA features are obtained in each scale, and the obtained multi-scale features can be combined to gain further improvement.

For multi-scale UA features, \ie, $\phi_\textrm{att}^{s1}, \phi_\textrm{att}^{s2}$, we first concatenate the two features $[\phi_\textrm{att}^{s1}; \phi_\textrm{att}^{s2}] \in \mathbb{R}^{2k}$, and then train a new project matrix $\mathbf{W}_\textrm{com} \in \mathbb{R}^{2k \times k}$ to obtain the combined UA feature, \ie, $\phi_\textrm{att}^\textrm{com} = \mathbf{W}_\textrm{com}^T [\phi_\textrm{att}^{s1}; \phi_\textrm{att}^{s2}]$.
For multi-scale LA features, \ie, $\phi_\textrm{lat}^{s1}, \phi_\textrm{lat}^{s2}$, the combined feature can be obtained by directly concatenating the normalized two features, \ie, $\phi_\textrm{lat}^\textrm{com}=[\widehat{\phi_\textrm{lat}^{s1}}; \widehat{\phi_\textrm{lat}^{s2}}]$. Finally, the ZSL prediction can be performed using (\ref{equ:prediction}) with the combined UA feature $\phi_\textrm{att}^\textrm{com}$ and the combined LA feature $\phi_\textrm{lat}^\textrm{com}$.

\section{Experiments}
\subsection{Datasets}
The proposed LDF model is evaluated on two representative ZSL benchmarks: Animals with Attributes (AwA) \cite{lampert2014attribute} and Caltech-UCSD Birds 200-2011 (CUB) \cite{wah2011caltech}. AwA includes 30,475 images from 50 common animals categories. The 85 class-level attributes (continuous) and the standard 40/10 zero-shot split are adopted in our experiments. The dataset of CUB is a fine-grained bird dataset with 200 different birds and 11,788 images. Following SynC \cite{changpinyo2016synthesized}, we use a split of 150/50 for zero-shot learning and utilize 312-dim attribute vectors at class level.
\subsection{Implementation Details}
\label{subsection:imple}
The FNets are initialized using two different CNN models pre-trained on ImageNet, \ie, GoogLeNet \cite{szegedy2015going} and VGG19 \cite{simonyan2014very} respectively, to learn, $\phi(x)$. For AwA, only one zoom operation is performed and the LDF model contains \textbf{two} scales, as the objects in AwA images are usually large and centered \footnote{In supplementary materials, we will show that if we use three scales on AwA, the third scale is actually \textbf{useless} for object recognition.}; for CUB, the LDF model includes \textbf{three} scales with two zoom-in operations (\ie, having two ZNets). In each scale, the size of each input image or zoomed region is 224$\times$224, following the same setting as the existing ZSL methods. During training, the LDF model is trained for 5 epoches for AwA and 20 epoches for CUB. The learning rates of GoogLeNet and VGG19 are \textit{fixed} and set to 0.0005 and 0.0001, respectively throughout all of the experiments. At the test stage, $\lambda$ in (\ref{equ:ridge}) is set to 1.0 for all datasets.

\textbf{Training strategy}: We first adopt the strategy used in \cite{fu2017look} to initial the ZNet. Then the other components in the LDF model are learned. The detailed process is as follows:

\textbf{Step 1}: The FNet in each scale is initialized with the \textit{same} GoogLeNet (or VGG19) pre-trained on ImageNet. Notice that in the subsequent steps of training, the parameters in each scale are \textit{not} shared.

\textbf{Step 2}: In each scale, the initialized FNet is utilized to search a discriminative square, which is then used to pre-train the ZNet. The size of the searched square is assumed to be the half size of the original image (\ie, $z_s=0.5$). Then we slide over the last convolutional layer in the FNet and select the region with the highest activations. Finally, the coordinates of the searched region ($[z_x, z_y, z_s]$) are utilized to train the zoom net with L2 loss.

\textbf{Step 3}: We keep the parameters of the ZNet fixed and train both the FNet and the ENet.

\textbf{Step 4}: Finally, the parameters of the whole LDF model are fine-tuned in an end-to-end approach.
\subsection{Baselines}
To verify the effectiveness of the different components in our LDF model, four baselines are designed to compare with the proposed LDF model.
\begin{itemize}
\item \textbf{SS-BE-Fixed} (Single Scale \& Baseline Embedding Model \& Fixed Image Representations). In this baseline, the ZNet is removed, and only the full-size images are utilized to extract image features. Moreover, the FNet is \textit{fixed} during the training. For semantic representations, only the user-defined attributes are considered (Section \ref{section:BE}).
    \vspace{-0.5em}
\item \textbf{SS-BE-Learned} (Single Scale \& Baseline Embedding Model \& Learned Image Representations). Compared with the SS-BE-Fixed baseline, the only difference is that the FNet \textit{can be learned} in this baseline.
    \vspace{-0.5em}
\item \textbf{SS-AE-Learned} (Single Scale \& Augmented Embedding Model \& Learned Image Representations). Compared with the SS-BE-Learned baseline, this baseline aims to build the \textit{augmented} embedding space (Section \ref{section:AE}) with considering both UA and LA.
    \vspace{-0.5em}
\item \textbf{MS-BE-Learned} (Multi Scale \& Baseline Embedding Model \& Learned Image Representations). Compared with the SS-BE-Learned baseline, the only difference is the ZNet is added into this model (Section \ref{section:ZNet}).
\end{itemize}

%
%
%

\begin{table}[t]
      \renewcommand{\arraystretch}{1.0}
      \caption{\label{tab:main}ZSL results (MCA, \%) on all the datasets using the deep features of VGG19 and GoogLeNet (numbers in parentheses). }
      \vspace{-1.5em}
      \begin{center}
      \begin{tabular}{|c|c|c|}
      \toprule
      \multicolumn{1}{c}{Method} & \multicolumn{1}{c}{AwA} & \multicolumn{1}{c}{CUB} \\ \midrule
      \multicolumn{1}{c}{DAP \cite{lampert2009learning}} & \multicolumn{1}{c}{57.2 (60.5)} & \multicolumn{1}{c}{44.5 (39.1)}  \\
      \multicolumn{1}{c}{ESZSL \cite{romera2015embarrassingly}} & \multicolumn{1}{c}{75.3 (59.6)} & \multicolumn{1}{c}{- (44.0)}  \\
      \multicolumn{1}{c}{SJE \cite{akata2015evaluation}}  & \multicolumn{1}{c}{- (66.7)} & \multicolumn{1}{c}{- (50.1)}  \\
      \multicolumn{1}{c}{LatEM \cite{xian2016latent}}  & \multicolumn{1}{c}{- (71.9)} & \multicolumn{1}{c}{- (45.5)}  \\
      \multicolumn{1}{c}{SynC \cite{changpinyo2016synthesized}} & \multicolumn{1}{c}{- (72.9)} & \multicolumn{1}{c}{- (54.5)}  \\
      \multicolumn{1}{c}{JLSE \cite{zhang2016zero}} & \multicolumn{1}{c}{80.46 (-)} & \multicolumn{1}{c}{42.11 (-)}  \\
      \multicolumn{1}{c}{MFMR \cite{xu2017matrix}} & \multicolumn{1}{c}{79.8 (76.6)} & \multicolumn{1}{c}{47.7 (46.2)}  \\
      \multicolumn{1}{c}{Low-Rank \cite{dinglow}} & \multicolumn{1}{c}{82.8 (76.6)} & \multicolumn{1}{c}{45.2 (56.2)}  \\
      \multicolumn{1}{c}{SCoRe \cite{morgado2017semantically}} & \multicolumn{1}{c}{82.8 (78.3)} & \multicolumn{1}{c}{59.5 (58.4)}  \\
      \multicolumn{1}{c}{LAD \cite{jiang2017learning}} & \multicolumn{1}{c}{82.48 (-)} & \multicolumn{1}{c}{56.63 (-)} \\
      \multicolumn{1}{c}{JSLA \cite{peng2017joint}} & \multicolumn{1}{c}{82.9 (-)} & \multicolumn{1}{c}{57.1 (-)}  \\ \midrule
      \multicolumn{1}{c}{SS-BE-Fixed (Ours)}  & \multicolumn{1}{c}{75.20 (73.70)} & \multicolumn{1}{c}{50.51 (50.31)}   \\
      \multicolumn{1}{c}{SS-BE-Learned (Ours)}  & \multicolumn{1}{c}{79.35 (75.19)} & \multicolumn{1}{c}{59.32 (58.26)}   \\
      \multicolumn{1}{c}{SS-AE-Learned (Ours)}  & \multicolumn{1}{c}{81.36 (77.77)} & \multicolumn{1}{c}{65.99 (66.96)}   \\
      \multicolumn{1}{c}{MS-BE-Learned (Ours)}  & \multicolumn{1}{c}{81.80 (78.31)} & \multicolumn{1}{c}{64.85 (64.39)}  \\
      \multicolumn{1}{c}{LDF (Ours)} & \multicolumn{1}{c}{\textbf{83.40} (79.13)} & \multicolumn{1}{c}{67.12 (\textbf{70.37})}   \\ \bottomrule
      \end{tabular}
      \footnotesize
      \end{center}
      \vspace{-2.5em}
      \end{table}

\subsection{Experimental Results}
The multi-way classification accuracy (MCA) is used for evaluating the ZSL models. The comparison results using two different CNN models are shown in Table \ref{tab:main}.

\noindent \textbf{Effect of feature learning.} From Table \ref{tab:main}, we first notice that, without any specially designed regularization terms, the SS-BE-Learned baseline has already achieved comparable performance with state-of-the-arts and marginally surpass the SS-BE-Fixed baseline. Most of existing ZSL methods use the fixed image feature and only focus on learning visual-semantic mapping with various human-designed regularization terms. We show that feature learning neglected in image feature extraction process is also important to ZSL, which should be paid more attentions. By simply fine-tuning the FNet in an end-to-end framework, SS-BE-Learned can make the image features associate with the semantic information of attributes for different ZSL tasks and obtain better performance.

\noindent \textbf{Effect of ZNet.} The MS-BE-Learned baseline aims to use the ZNet to automatically discover discriminative regions from full-size images and leverage the coarse-to-fine representations to obtain better performance. We can see that the performance of MS-BE-Learned baseline outperforms both the SE-BE-Learned baseline and most of the state-of-the-art methods (Table \ref{tab:main}, 81.80\% on AwA, 64.85\% on CUB).

We further analyze the performance of each scale in MS-BE-Learned model, and show the results in Table \ref{tab:multiscale}. It can be seen that, the performance of the first scale, \ie, MS-BE-Learned (Scale 1), is comparable with the single scale baseline, SS-BE-Learned. With more discriminative image features utilized, the performance of the second and the third scale improves continuously.


\noindent \textbf{Effect of the latent attribute modelling.} The SS-AE-Learned baseline aims to build an augmented embedding space. It is more reasonable to associate image features with both user-defined and latent attributes in our augmented space. It can be observed from Table \ref{tab:main} that the SS-AE-Learned model outperforms SE-BE-Learned baseline for both AwA (81.36\%) and CUB (66.96\%) datasets.

\begin{table}[t]
      \renewcommand{\arraystretch}{1.0}
      \small
      \caption{\label{tab:multiscale}The detailed ZSL results (\%) on each scale. }
      \begin{center}
      \begin{tabular}{|c|c|c|}
      \toprule
      \multicolumn{1}{c}{Method}  & \multicolumn{1}{c}{AwA} & \multicolumn{1}{c}{CUB} \\ \midrule
      \multicolumn{1}{c}{SS-BE-Learned } & \multicolumn{1}{c}{79.35 (75.19)} & \multicolumn{1}{c}{59.32 (58.26)}  \\ \midrule
      \multicolumn{1}{c}{MS-BE-Learned (Scale 1)} & \multicolumn{1}{c}{79.20 (75.68)}  & \multicolumn{1}{c}{59.88 (58.87)}  \\
      \multicolumn{1}{c}{MS-BE-Learned (Scale 2)} & \multicolumn{1}{c}{79.87 (77.02)}  & \multicolumn{1}{c}{61.04 (61.81)}  \\
      \multicolumn{1}{c}{MS-BE-Learned (Scale 3)} & \multicolumn{1}{c}{- (-)}  & \multicolumn{1}{c}{62.04 (62.72)}  \\ \midrule
      \multicolumn{1}{c}{MS-BE-Learned (All Scale)} & \multicolumn{1}{c}{81.80 (78.31)}  & \multicolumn{1}{c}{64.85 (64.39)}  \\ \bottomrule
      \end{tabular}
      \footnotesize
      \begin{tablenotes}
      \item [1] MS-BE-Learned (Scale X) denotes the ZSL results using the image features of scale X only.
      \end{tablenotes}
      \end{center}
      \vspace{-1.5em}
      \end{table}

      \begin{table}[t]
      \renewcommand{\arraystretch}{1.0}
      \small
      \caption{\label{tab:UA}ZSL results (\%) with UA features or LA features only. }
      \vspace{-1em}
      \begin{center}
      \begin{tabular}{|c|c|c|}
      \toprule
      \multicolumn{1}{c}{Method}  & \multicolumn{1}{c}{AwA} & \multicolumn{1}{c}{CUB} \\ \midrule
      \multicolumn{1}{c}{SS-BE-Learned }  & \multicolumn{1}{c}{79.35 (75.19)} & \multicolumn{1}{c}{59.32 (58.26)} \\ \midrule
      \multicolumn{1}{c}{SS-AE-Learned (UA)}   & \multicolumn{1}{c}{80.97 (77.24)} & \multicolumn{1}{c}{62.17 (59.40)} \\  \midrule
      \multicolumn{1}{c}{SS-AE-Learned (LA)}   & \multicolumn{1}{c}{78.76 (75.75)} & \multicolumn{1}{c}{63.08 (66.11)} \\
      \multicolumn{1}{c}{SS-AE-Learned (UA \& LA)}   & \multicolumn{1}{c}{81.36 (77.77)} & \multicolumn{1}{c}{65.99 (66.96)} \\   \bottomrule
      \end{tabular}
      \footnotesize
      \begin{tablenotes}
      \item [1] SS-AE-Learned (UA/LA) denotes the results predicted with the UA features $\phi_\textrm{att}(x)$ only or the LA features $\phi_\textrm{lat}(x)$ only.
      \end{tablenotes}
      \end{center}
      \vspace{-2.5em}
      \end{table}

We believe that, in the augmented attribute space, the learning of LA will help the learning of UA. Further experiments are conducted to verify this. The results are shown in Table \ref{tab:UA}. For SS-AE-Learned baseline, we only utilize the obtained UA representation $\phi_{\textrm{att}}(x)$ to perform ZSL prediction as in (\ref{equ:UA}), denoted as SS-AE-Learned (UA). We can see that, when using UA features only, the performance of SS-AE-Learned (UA) is higher than the SS-BE-Learned. (\eg, 80.97\% \vs 79.35\%). It proves that better UA representations are obtained in the augmented attribute space.

\noindent \textbf{Comparisons with state-of-the-art methods.} Compared with previous methods in Table \ref{tab:main}, the LDF model improves the state-of-the-art performance on both datasets. In general, the proposed model based on VGG19 performs better on AwA, while the GoogLeNet-based model shows superiority on CUB. On AwA, our LDF achieves 83.40\%, which is slightly higher than JLSA \cite{peng2017joint} (82.81\%). For more challenging CUB dataset that 50 bird species need to be classified, our model obtains more obvious improvement. On CUB, the LDF model reaches 70.37\%, with an impressive gain over the state-of-the-art SCoRe (from 58.4\% to 70.37\%).

Furthermore, the components of the latent discriminative regions mining (the ZNet) and the latent discriminative attribute modelling (the ENet) are jointly learned in the proposed LDF model. We believe the two components could assist each other in the joint learning framework. To verity this assumption, a further analysis of the LDF model is performed, and the results are shown in Table \ref{tab:joint}. It can be seen that, when using the combined UA features only to perform ZSL prediction, \ie, LDF (UA), the performance of LDF is higher than the MS-BE-Learned baseline. When using the combined LA features only, the performance of the LDF (LA) also exceeds the SS-AE-Learned (LA). It confirms the advantages of the jointly learning approach.

 \begin{table}[t]
      \renewcommand{\arraystretch}{1.0}
      \small
      \caption{\label{tab:joint}The comparisons between the joint training and separated training for ZNet and ENet. }
      \vspace{-1em}
      \begin{center}
      \begin{tabular}{|c|c|c|}
      \toprule
      \multicolumn{1}{c}{Method} & \multicolumn{1}{c}{AwA} & \multicolumn{1}{c}{CUB} \\ \midrule
      \multicolumn{1}{c}{SS-AE-Learned (LA) }  & \multicolumn{1}{c}{78.76 (75.75)} & \multicolumn{1}{c}{63.08 (66.11)}  \\
      \multicolumn{1}{c}{LDF (LA)} & \multicolumn{1}{c}{79.35 (76.84)}  & \multicolumn{1}{c}{66.47 (69.94)}  \\ \midrule
      \multicolumn{1}{c}{MS-BE-Learned (UA)} & \multicolumn{1}{c}{81.80 (78.31)}  & \multicolumn{1}{c}{64.85 (64.39)} \\
      \multicolumn{1}{c}{LDF (UA)} & \multicolumn{1}{c}{82.47 (78.77)}  & \multicolumn{1}{c}{65.94 (65.78)}  \\ \midrule
      \multicolumn{1}{c}{LDF (LA \& UA)} & \multicolumn{1}{c}{83.40 (79.13)}  & \multicolumn{1}{c}{67.12 (70.37)}  \\ \bottomrule
      \end{tabular}
      \footnotesize
      \begin{tablenotes}
      \item [1]  LDF (LA/UA) denotes the ZSL results predicted with the combined LA features $\phi_\textrm{lat}^{\textrm{com}}$ only or the combined UA features $\phi_\textrm{att}^{\textrm{com}}$ only.
      \end{tablenotes}
      \end{center}
      \vspace{-1.5em}
      \end{table}

\begin{figure}[]
    \begin{center}
    \advance\leftskip -0.5cm
        \includegraphics[height=3.8cm]{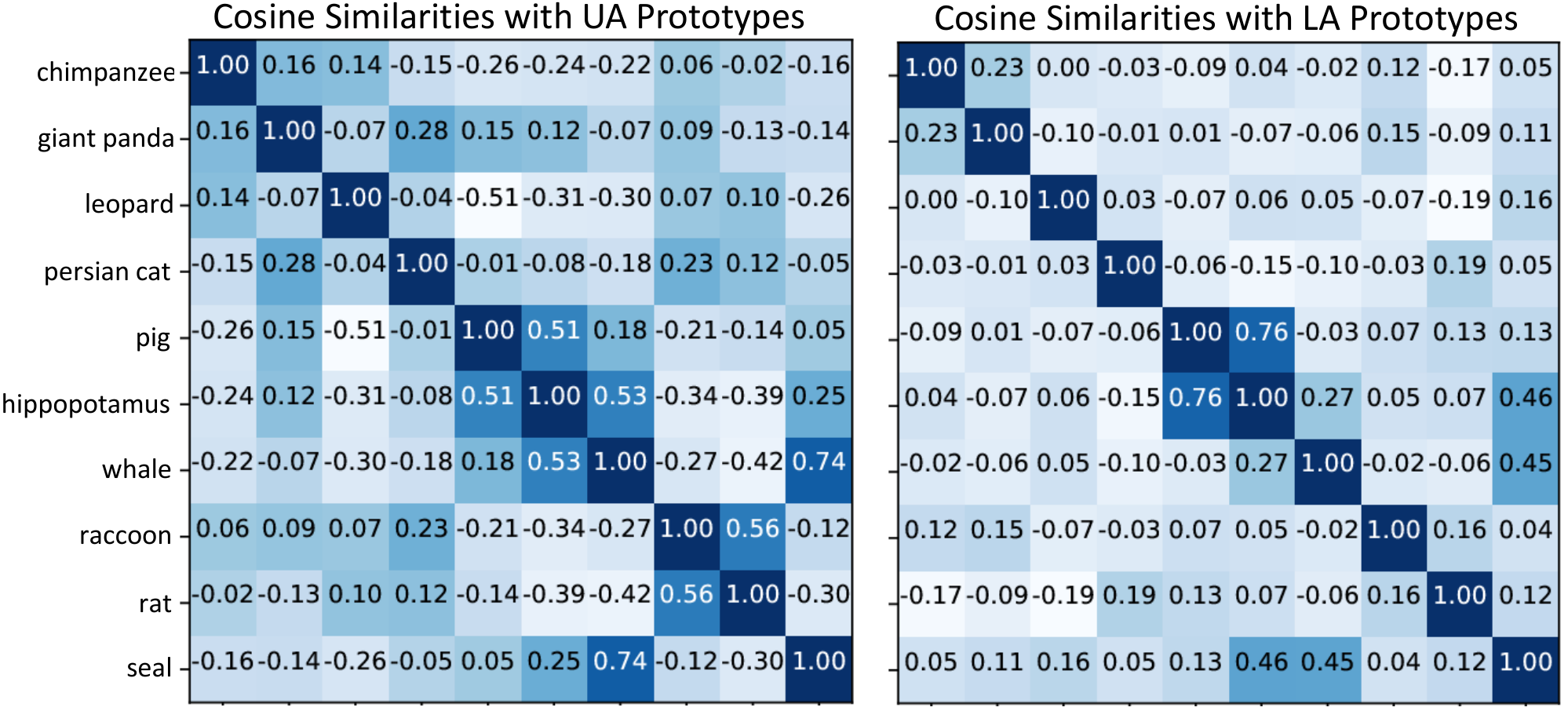}
    \end{center}
    \vspace{-1em}
        \caption{The cosine similarities computed with the UA (left panel) and the LA (right panel) for 10 unseen AwA classes.}
    \label{fig:cosine}
    \vspace{-1em}
    \end{figure}

\noindent \textbf{Discriminativeness of LA.} The LA features are learned to be discriminative by exploiting the category information as in (\ref{equ:lat_loss}), and we believe the learned LA space is more discriminative than the UA space. To illustrate this, we show some examples on AwA in Figure \ref{fig:visual}. The test images are projected to their UA features and LA features with (\ref{equ:two_proj}). Then for a UA element or a LA element, the images which have largest and smallest activations of the component are shown. It can be observed that, for LA features, the images with large activations belong to one same category and the images with small activations are of the other category. In contrast, the user-defined attributes are usually shared in multiple categories. It confirms the apparent discriminative property of the learned latent attributes.

\begin{figure*}[t]
    \begin{center}
        \includegraphics[height=6.6cm]{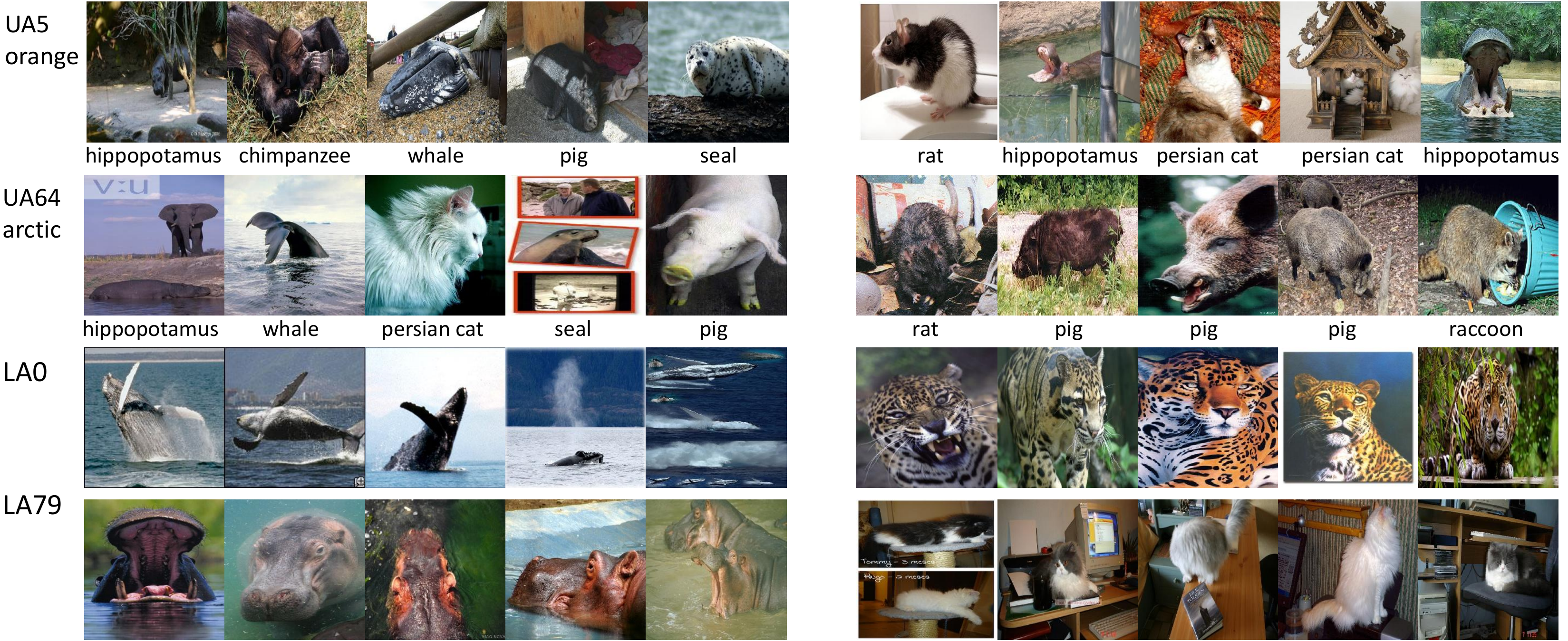}
    \end{center}
    \vspace{-1em}
        \caption{The visual examples on AwA with VGG19 SS-AE-Learned. `UA/LAX' denotes the X-th element of the attribute features. In each row, the first five images are top-5 images with largest activations and the last five images are selected images with smallest activations.}
    \label{fig:visual}
    \vspace{-1em}
    \end{figure*}

Additionally, to quantitatively compare the learned LA space with the UA space, we calculate cosine similarities between unseen classes with both the LA and UA prototypes, and the results are shown in Figure \ref{fig:cosine}. The LA prototypes are obtained by directly averaging the LA features, \ie, $\overline{\phi_\textrm{lat}} = \frac{1}{N} \sum_i \phi_\textrm{lat}(x_i)$, for each unseen class, and the UA prototypes are the class-level attribute annotations, \ie, $\mathbf{a}^c$. It can be seen that, compared with the UA prototypes, the cosine similarities between different LA prototypes are obviously smaller for most categories, except for the \textit{pig} and the \textit{hippopotamus}. Compared with attributes annotated by experts, our LA prototypes are learned from the images only. Thus, the categories with similar appearances, \eg, \textit{pig} \vs \textit{hippopotamus}, get closer in the LA space.

It is noted that when we perform ZSL prediction with LA features, a LA representation (prototype) of a test \textit{category} is needed, but absent in the dataset. Thus, the LA prototypes for unseen classes have to be computed with (\ref{equ:lat_prototype}) leveraging the relationship $\beta_c$. However, $\beta_c$ is computed in the UA space and it cannot exactly reflect the true relationship between LA prototypes. This bias finally degrades the ZSL performance when LA prototypes are utilized for prediction with (\ref{equ:lat_prediction}). This bias explains why, in Table \ref{tab:UA}, the performance of SS-AE-Learned (LA) is lower than SS-AE-Learned (UA) on AwA, although the learned LA space is actually more discriminative than the UA space.

\begin{figure}[]
    \begin{center}
        \includegraphics[width=7.5cm]{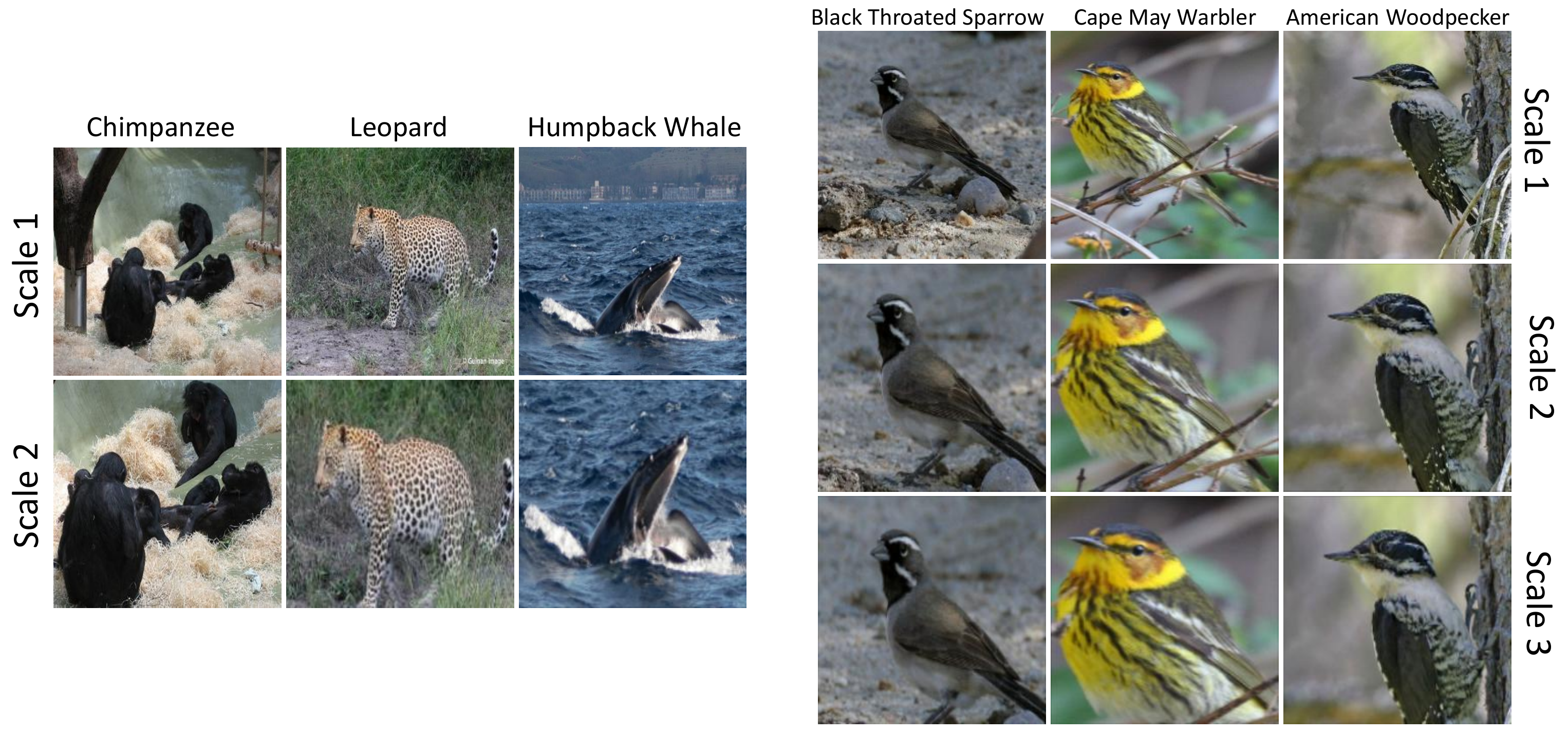}
    \end{center}
    \vspace{-1em}
        \caption{The examples of the learned regions at different scales.}
    \label{fig:cub}
    \vspace{-1.5em}
    \end{figure}

\noindent \textbf{Visualizations of discriminative regions.} In Figure \ref{fig:cub}, we show the discovered regions with the LDF model. The left three columns show the examples selected from AwA. We can see that, for images with a single instance, the LDF model progressively searches for finer regions until it finds the main object; for images with multiple instances, the model tends to find a large square including the multiple objects. Another interesting discovery on AwA is that, for some specific categories, \eg, \textit{whale}, the identified regions will include obvious more background elements than others. The reason is that the searched regions of the \textit{humpback whale} are required to be matched with their user-defined attributes, some of which, such as \textit{swims}, \textit{water} and \textit{ocean}, highly relate to the background waters in the images.

The examples in right three columns are sampled from CUB. It is aware that the CUB dataset provides bounding box annotations, however, our model could automatically discover object-centric regions without such annotations, which shows another advantage of our framework. It is noted that, the network in \cite{fu2017look} performs fine-grained object recognition, a different task from us; and it could discover some object parts. In contrast, in our ZSL model, the searched regions should be associated to the user-defined attributes, which, for example, correspond to the whole body of the birds from bills to tails. Thus, it is expected that the model will focus on regions containing the whole object rather than its parts; and our analysis confirms this.


\section{Conclusion}
In this paper, an end-to-end model is proposed to learn the latent discriminative features for ZSL in both visual and semantic space. For visual space, we introduce the zoom net to automatically search for discriminative regions. For semantic space, we propose an augmented attribute space with both the user-defined attributes and the latent attributes. The latent attributes are learned to be discriminative with category information. Finally, the two components could assist each other in the end-to-end joint learning framework.

\section{Acknowledgement}
This work is funded by the National Key Research and Development Program of China (Grant 2016YFB1001004 and Grant 2016YFB1001005), the National Natural Science Foundation of China (Grant 61673375, Grant 61721004 and Grant 61403383) and the Projects of Chinese Academy of Sciences (Grant QYZDB-SSW-JSC006 and Grant 173211KYSB20160008).

{\small
\bibliographystyle{ieee}
\bibliography{egbib}
}
\clearpage
\begin{appendices}
\section{How to Identify the Discriminative Region from an Image?}
To search the discriminative region from an image in zero-shot learning (ZSL), two weakly supervised learning approaches can be considered: 1) directly regressing the locations of the identified region (\eg, the proposed zoom scheme in our LDF model); 2) extracting multiple region proposals (\eg, EdgeBox \cite{zitnick2014edge}) for the image and then selecting the most discriminative one. In this paper, we didn't utilize the latter region proposal method based on the following considerations. First, the goal of the region proposal algorithm \cite{zitnick2014edge} is to identify ``objects''. However, as shown in Figure \ref{fig:cub} and claimed in Section \ref{section:ZNet}, in ZSL, the identified region may contain context elements to match its user-defined attributes. Such region is not exactly equal to the ``object'' region and hard to be captured by EdgeBox. Second, processing multiple proposals (typically 2,000) for each image is quite inefficient, and selecting the proper region from 2,000 ones is also difficult in weakly supervised settings. We have conducted an experiment to test the region proposal approach for ZSL.

Specially, we first extract 2,000 EdgeBox proposals for each image. Then we replace the \textit{pool5} layer in SS-BE-baseline (VGG19) with the RoI Pooling layer proposed in Fast RCNN \cite{girshick2015fast}. The images with their region proposals are imported into the model, and the model could output the compatibility score for each region. Following the standard multiple instance learning (MIL) setting, the region with highest compatibility score is selected to compute the loss function as in (\ref{equ:baseline}). The network finally obtains 72.67\% on AwA dataset. This result is even lower than SS-BE-Learned (Table \ref{tab:main}, 78.35\%), which directly extract image features from full-size images. Moreover, the runtime is 7$\sim$8 times longer than our zoom scheme.

\section{The Bilinear Interpolation Operation}
In Section \ref{section:ZNet}, to obtain better representation for finer localized cropped region $x^{\text{crop}}$, the bilinear interpolation is utilized to adaptively zoom the cropped region to the same size with the original image. Concretely, for a point $(i,j)$ of the zoomed region, its value $x^{\text{zoom}}_{(i,j)}$ can be computed by linearly combining the values of nearest four points in the cropped region. Formally,
\begin{equation}
\begin{aligned}
 x^{\text{zoom}}_{(i,j)} = & \sum_{\alpha,\beta} |1-\alpha-\{i/\lambda\}| |1-\beta-\{j/\lambda\}|x^{\textrm{crop}}_{(m,n)}, \\
 m = & \quad [i/\lambda]+\alpha+z_x-z_s, \quad \alpha = 0,1\\
 n = & \quad [j/\lambda]+\beta+z_y-z_s, \quad \beta = 0,1
\end{aligned}
\end{equation}

where $\lambda$ is the upsampling factor, \ie, $\lambda=1/z_s$. $[\cdot]$ and $\{\cdot\}$ is the integral and fractional part, respectively.

\section{Experiments with Three Scales on AwA}

As we have mentioned in Section \ref{subsection:imple}, for AwA dataset, only one zoom operation is performed and the two-scale model is adopted. We claim the reason is that the objects in AwA images are usually large and centered. To verify this, in this section, we analyze the performance of three-scale MS-BE-Learned baseline on AwA. The experiment is conducted with GoogLeNet and all the experimental settings are the same as we described in Section \ref{subsection:imple}. The performance of each single scale is shown in Table \ref{tab:multiscaleawa}.

Additionally, the parameter $z_s$ in (\ref{equ:znet}) represents the length of the cropped regions. In scale 1 and scale 2, we respectively count the $z_s$ values for all unseen images and show the mean value of the $z_s$ in Table \ref{tab:multiscaleawa}. It can be seen that when the three-scale model is adopted on AwA, the performance of the second scale is higher than the first scale (77.12\% \vs 75.47\%). However, the performance of the third scale does not show the further improvement (77.05\% \vs 77.12\%). When we inspect the mean $z_s$ values in the second scale, it can be found that the scale size of the cropped region is nearly 1 (0.98), that is, the zoom net in the second scale actually does not perform any cropping operation and directly send the original image to the third scale. As we have claimed, the objects in AwA images are large and centered. Through one time zoom operation, the network can capture the main object and the third scale is actually useless in the model.

\begin{table}[t]
      \renewcommand{\arraystretch}{1.0}
      \scriptsize
      \begin{center}
      \begin{tabular}{|c|c|c|}
      \toprule
      \multicolumn{1}{c}{}  & \multicolumn{1}{c}{ZSL performance on AwA} & \multicolumn{1}{c}{mean value of $z_s$} \\ \midrule
      \multicolumn{1}{c}{SS-BE-Learned } & \multicolumn{1}{c}{75.19} & \multicolumn{1}{c}{-}  \\ \midrule
      \multicolumn{1}{c}{MS-BE-Learned (Scale 1)} & \multicolumn{1}{c}{75.47}  & \multicolumn{1}{c}{0.87}  \\
      \multicolumn{1}{c}{MS-BE-Learned (Scale 2)} & \multicolumn{1}{c}{77.12}  & \multicolumn{1}{c}{0.98}  \\ \midrule
      \multicolumn{1}{c}{MS-BE-Learned (Scale 3)} & \multicolumn{1}{c}{77.05}  & \multicolumn{1}{c}{-}  \\ \bottomrule
      \end{tabular}
      \end{center}
            \caption{\label{tab:multiscaleawa}The detailed ZSL results (\%) on each scale and the mean value of $z_s$ parameter. }
      \end{table}

\begin{table*}[t]
      \renewcommand{\arraystretch}{1.0}
      \begin{center}
      \begin{tabular}{|c|c|c|}
      \toprule
      \multicolumn{1}{c}{}  & \multicolumn{1}{c}{ZSL performance on AwA}  & \multicolumn{1}{c}{The dimension of LA} \\ \midrule
      \multicolumn{1}{c}{SS-AE-Learned (LA) } & \multicolumn{1}{c}{75.75}  & \multicolumn{1}{c}{$k$ (85)}\\
      \multicolumn{1}{c}{SS-AE-Learned (LA)} & \multicolumn{1}{c}{75.83}   & \multicolumn{1}{c}{$2k$ (170)} \\
      \multicolumn{1}{c}{SS-AE-Learned (LA)} & \multicolumn{1}{c}{76.01}   & \multicolumn{1}{c}{$3k$ (255)} \\ \bottomrule
      \end{tabular}
      \end{center}
      \caption{\label{tab:dim}The ZSL results (\%) with different dimension of latent attributes. }
      \end{table*}

\section{The Effect of the Dimension of Latent Attribute}
As we mentioned in Section \ref{section:AE}, the dimension of the latent attributes (LA) is set to $k$, \ie, the same with the user-defined attributes (UA). In this section, we explore the effectiveness of the latent attributes' dimension and conduct experiments on AwA dataset with GoogLeNet. Specially, we train the SS-AE-Learned baseline with different dimensions of LA (\ie, $k$, $2k$ and $3k$), and perform ZSL prediction with the latent attributes only. The results are shown in Table \ref{tab:dim}. It can be seen that with the larger dimension of LA, the ZSL performance improves. But the improvement is slight and the performance in general is robust to the dimension of LA.

\section{The Discriminativeness of the Learned Latent Attributes}
\begin{figure*}[]
    \begin{center}
        \includegraphics[height=13.5cm]{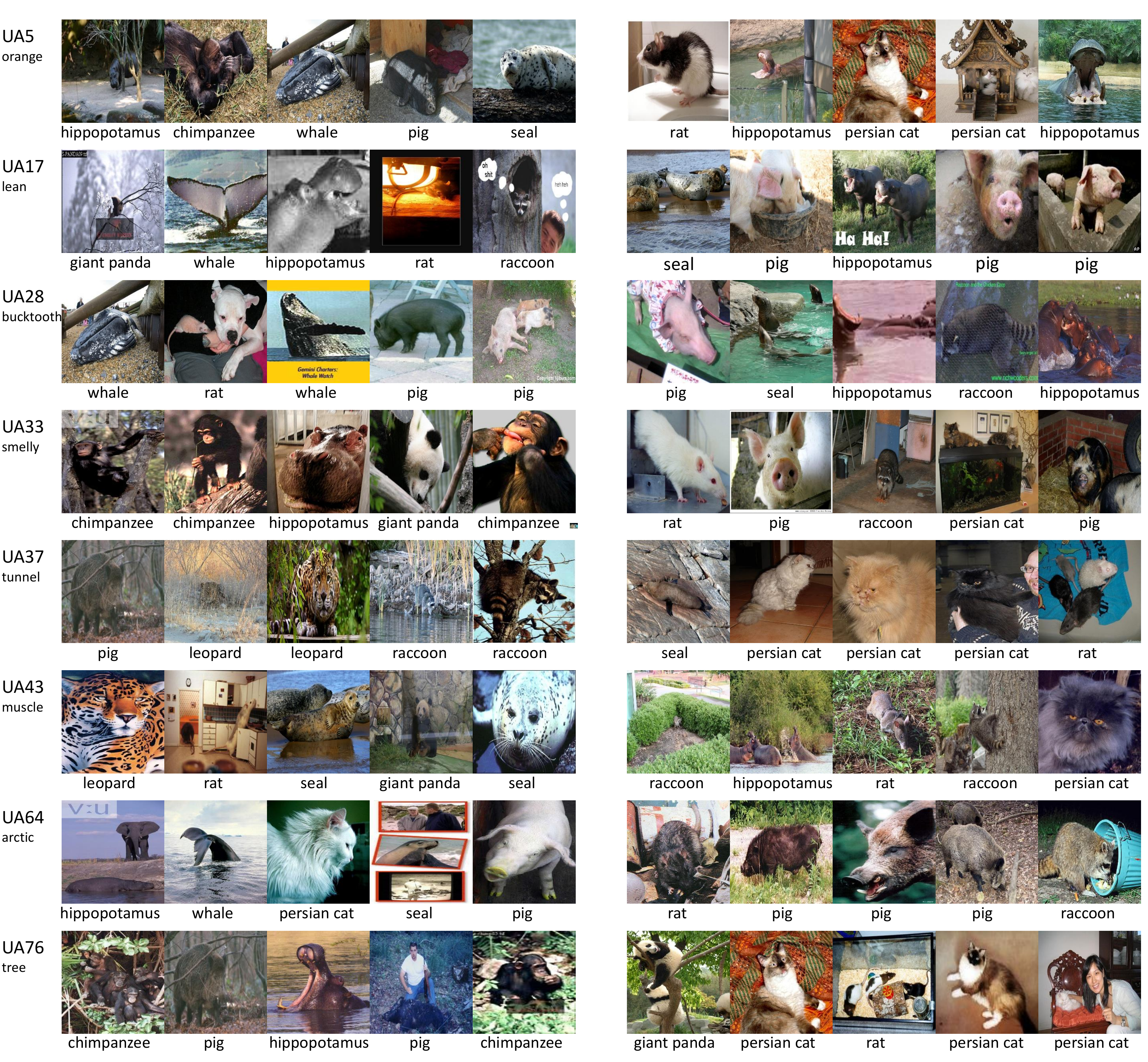}
    \end{center}
        \caption{The visual examples of user-defined attributes (UA) on AwA. `UDAX' denotes the X-th element of the UA features. The UA features are obtained with the VGG19 SS-AE-Learned baseline. The first five images are top-5 images with largest activations over this UA element and the last five images are selected examples with smallest activations.}
    \label{fig:visual-uda}
    \end{figure*}

In this section, we show more visualized examples to illustrate the discriminative property of latent attributes. For a latent attribute element, the images which have largest and smallest activations over this element are shown in Figure \ref{fig:visual}. Meanwhile, the examples selected with the learned UA features are shown in Figure \ref{fig:visual-uda} for comparison. From Figure \ref{fig:visual-uda}, it can be seen that the user-defined attributes are shared in many objects. Another discovery is that the prediction results of user-defined attributes will be affected by mid-level cues, \eg, colors. For example, for UDA5 element, the \textit{chimpanzee}, \textit{whale} and \textit{pig} objects are falsely predicted as \textit{orange} due to the existing orange backgrounds. For UDA64 element, the \textit{persian cat} and \textit{pig} images are falsely predicted as \textit{arctic}. It is possible that the two animals share white appearances.

\begin{table*}[]
      \renewcommand{\arraystretch}{1.0}
      \begin{center}
      \begin{tabular}{ccccccc}
      \toprule
                                  & \multicolumn{3}{c}{AwA} & \multicolumn{3}{c}{CUB} \\
      \multicolumn{1}{c}{Method}  & \multicolumn{1}{c}{$A_{\mathcal{U}\rightarrow\mathcal{T}}$} & $A_{\mathcal{S}\rightarrow\mathcal{T}}$ & $H$ & \multicolumn{1}{c}{$A_{\mathcal{U}\rightarrow\mathcal{T}}$} & $A_{\mathcal{S}\rightarrow\mathcal{T}}$ & $H$\\ \midrule

      \multicolumn{1}{c}{DAP \cite{lampert2014attribute}$^*$ }  & 2.4 & 77.9 & 4.7 & 4.0 & 55.1 & 7.5 \\
      \multicolumn{1}{c}{IAP \cite{lampert2014attribute}$^*$}   & 1.7 & 76.8 & 3.3 & 1.0 & 69.4 & 2.0 \\
      \multicolumn{1}{c}{ConSE \cite{norouzi2013zero}$^*$} & 9.5 & 75.9 & 16.9 & 1.8 & 69.9 & 3.5 \\
      \multicolumn{1}{c}{SynC$^\text{o-vs-o}$ \cite{changpinyo2016synthesized}$^*$}    & 0.3 & 67.3 & 0.6 & 8.4 & 66.5 & 14.9 \\
      \multicolumn{1}{c}{SynC$^\text{struct}$ \cite{changpinyo2016synthesized}$^*$}   & 0.4 & 81.0 & 0.8 & 13.2 & 72.0 & 22.3 \\ \midrule
      \multicolumn{1}{c}{LDF (Ours)}   & \textbf{9.8} & \textbf{87.4} & \textbf{17.6} & \textbf{26.4} & \textbf{81.6} & \textbf{39.9} \\\bottomrule
      \end{tabular}
      \end{center}
            \caption{\label{tab:gzsl}Generalized zero-shot learning results (\%). All results are obtained with GoogLeNet features. $*$ means that the numbers of the method are cited from \cite{chao2016empirical}, since the original paper does not report the gZSL results. $H$ denotes the harmonic mean. }
      \end{table*}

\section{Generalized Zero-Shot Learning Results}
In conventional zero-shot learning (cZSL), ZSL methods are trained on seen classes and evaluated on unseen ones. The basic assumption in cZSL is that test instances always come from the unseen classes (denoted as $\mathcal{U}\rightarrow\mathcal{U}$), which is actually unrealistic in real-world applications. Motivated by this, recent ZSL works \cite{chao2016empirical,xian2017zero} aim to measure the zero-shot performance in the generalized zero-shot learning (gZSL) setting. In gZSL, the test images are assumed to come from all target classes including both seen and unseen categories.

Similar to \cite{chao2016empirical}, 20\% of the images from seen classes are extracted and then merged with the images from unseen classes to form the new test set. We denoted the joint label space of seen and unseen classes as $\mathcal{T} = \mathcal{S} \cup \mathcal{U}$ and evaluate the proposed LDF model in terms of accuracy on $\mathcal{U}\rightarrow\mathcal{T}$ and $\mathcal{S}\rightarrow\mathcal{T}$, which are denoted as $A_{\mathcal{U}\rightarrow\mathcal{T}}$ and $A_{\mathcal{S}\rightarrow\mathcal{T}}$, respectively. $A_{\mathcal{U}\rightarrow\mathcal{T}}$ indicates the accuracies of classifying test images from unseen classes into the joint label space while $A_{\mathcal{S}\rightarrow\mathcal{T}}$ indicates the accuracies of recognizing seen objects into the joint label space. Moreover, similar to \cite{xian2017zero}, the harmonic mean is computed to measure the ZSL methods with considering both the accuracy of seen classes and the accuracy of unseen classes. Formally,

\begin{equation}
H = \frac {2A_{\mathcal{U}\rightarrow\mathcal{T}}A_{\mathcal{S}\rightarrow\mathcal{T}}} {A_{\mathcal{U}\rightarrow\mathcal{T}} + A_{\mathcal{S}\rightarrow\mathcal{T}}}
\end{equation}

The experiments are performed on both AwA and CUB datasets. The GoogLeNet model is utilized and the results are shown in Table \ref{tab:gzsl}. It can be seen that on both datasets, the proposed LDF model significantly outperforms previous methods on all the three metrics, which confirms the advantage of our method under the gZSL setting.

\newpage
\begin{figure*}[]
    \begin{center}
        \includegraphics[height=13.5cm]{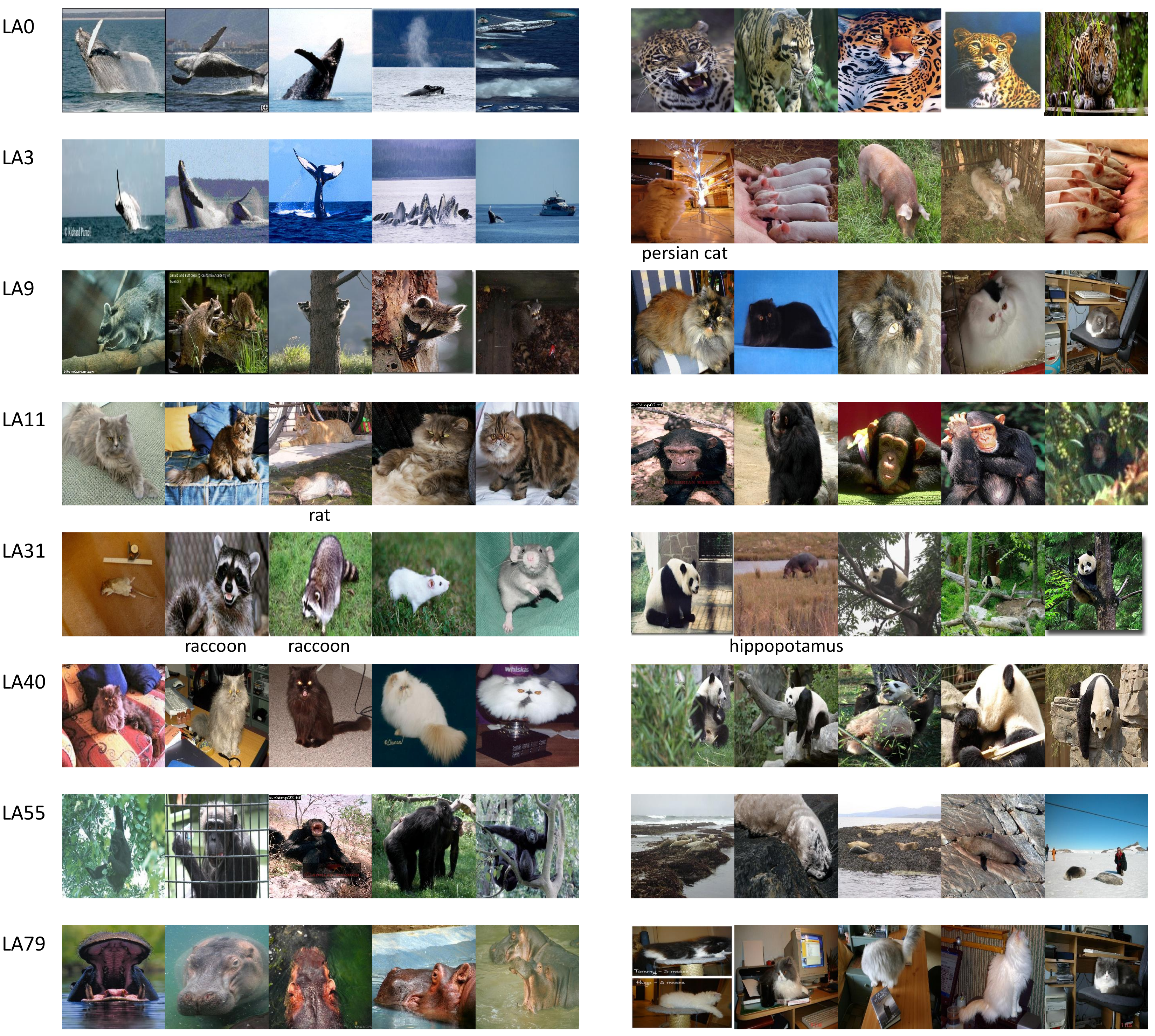}
    \end{center}
        \caption{The visual examples of latent discriminative attributes (LA) on AwA. `LAX' denotes the X-th element of the LA features. The LA features are obtained with the VGG19 SS-AE-Learned baseline. The first five images are top-5 images with largest activations over this element and the last five images are selected examples with smallest activations.}
    \label{fig:visual}
    \end{figure*}

\end{appendices}

\end{document}